\documentclass[letterpaper, 10 pt, conference]{ieeeconf}  %

\IEEEoverridecommandlockouts                              %

\usepackage{graphics} %
\usepackage{graphicx}
\usepackage{subcaption}

\usepackage{bm}

\usepackage[noadjust]{cite}

\usepackage{hyperref}
\usepackage{amsmath} %
\usepackage{amssymb}  %
\usepackage[ruled,linesnumbered,vlined]{algorithm2e}
\usepackage{booktabs}
\usepackage{multirow}
\usepackage{cleveref}
\usepackage{color, soul}
\usepackage{array}
\newcolumntype{C}[1]{>{\centering\arraybackslash}p{#1}}

\title{\LARGE \bf
  Monocular Direct Sparse Localization in a Prior 3D Surfel Map
}

\def\thankstext{This work was supported in part by the National Natural Science Foundation of China (U1713211), the Guangdong-Hong Kong Cooperation Innovation Platform (Grant No. 2018B050502009), and the Research Grant Council of Hong Kong SAR Government, China (No. 11210017 and No. 21202816), awarded to Prof. Ming Liu. \par
The authors are with \href{https://ram-lab.com/}{RAM-LAB}, the Hong Kong University of Science and Technology, Kowloon, Hong Kong {\tt\small hy.ye@connect.ust.hk}, {\tt\small hhuangat@connect.ust.hk}, {\tt\small eelium@ust.hk}}

\author{Haoyang~Ye, Huaiyang~Huang and~Ming~Liu%
\thanks{\thankstext}%
}

\begin{document}

\maketitle
\thispagestyle{empty}
\pagestyle{empty}

\begin{abstract}
  In this paper, we introduce an approach to tracking the pose of a monocular camera in a prior surfel map.
  By rendering vertex and normal maps from the prior surfel map, the global planar information for the sparse tracked points in the image frame is obtained.
  The tracked points with and without the global planar information involve both global and local constraints of frames to the system.
  Our approach formulates all constraints in the form of direct photometric errors within a local window of the frames.
  The final optimization utilizes these constraints to provide the accurate estimation of global 6-DoF camera poses with the absolute scale.
  The extensive simulation and real-world experiments demonstrate that our monocular method can provide accurate camera localization results under various conditions.
\end{abstract}

\setlength{\textfloatsep}{0.5pt}

\section{Introduction}

Localization is one of the fundamental requirements for autonomous vehicles.
Various sensors and algorithms have been developed to fulfill real-time localization or simultaneous localization and mapping (SLAM).
Though GPS can provide global position information over the earth, the localization results can be easily influenced by multi-path effects and it cannot be used indoors.

On-board sensors which do not rely on extrinsic infrastructure have become a necessity for reliable localization.
Cameras and lidars are two of the most popular sensors employed for the tasks of localization and SLAM.
Lidars can provide accurate and long-range measurements of the environment, and many lidar-based localization and mapping methods \cite{zhang2014loam, shan2018lego, ye2019tightly} show good performance indoors and outdoors.
However, a typical 3D lidar is bulky and expensive, which limits its application on some small or low-cost platforms.

Cameras have become an alternative to lidars thanks to their light weight and low cost.
Camera-based methods or those fused with an inertial measurement unit (IMU), visual-inertial methods, can meet the same demands for localization \cite{mur2015orb,engel2017direct,qin2018vins}.
However, compared to lidar-based systems, they have inferior accuracy and robustness.
In particular, monocular camera systems will face the scale drift problem \cite{strasdat2010scale}.
Appearance changes, including weather and illumination, can also cause instability to camera-based methods.
Finding associations in multiple session maps can address the problem \cite{churchill2013experience}, but it is costly to store multiple maps of the same place.

An affordable way to combine the advantages of lidars and cameras is to use a 3D lidar to build a 3D map and then achieve camera-based localization in this built map.
In this way, accurate and large-scale maps can be efficiently built by 3D lidar mapping methods.
Then, the cameras can utilize the geometric information from the map to reduce the long-term drift and gain more accurate localization results.

\begin{figure}[!t]
  \centering
  \includegraphics[width=0.48\textwidth]{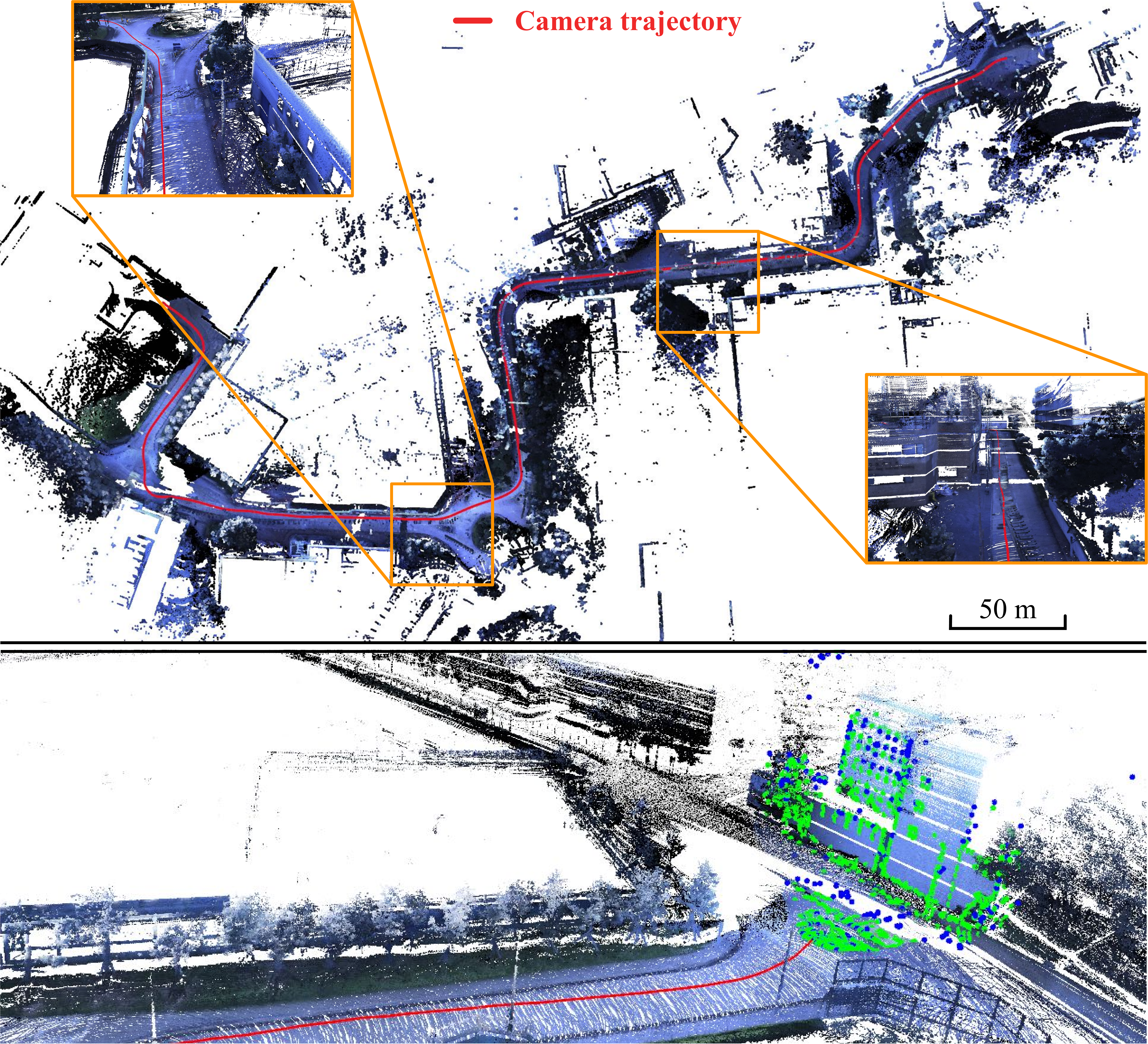}
  \caption{Our method localizes a monocular camera in the surfel map.
    The red trajectory shows the camera positions in the 3D map, which is colorized by the color information from the camera (top).
    Green points are those with surfel constraints, while blue points are those without surfel constraints in a local window of frames (bottom).}
  \label{fig:to_ias}
\end{figure}

Based on this idea, we present a novel monocular camera localization system in a 3D surfel map, called DSL (Direct Sparse Localization).
The main contributions of our paper are as follows,
\begin{itemize}
  \item A cross-modality localization algorithm is proposed to localize camera poses in a prior surfel map.
        All the constraints are from direct photometric energy functions, making our system efficient to track and optimize camera poses.

  \item Global constraints from the map make the monocular system aware of the absolute scale and global transform. We adopt the surfel representation, making our method efficient to store 3D information and render depth, along with vertex \& normal maps with a modern GPU.

  \item Degeneration analysis of our method, which can provide a hint as to the uncertainty of localization accuracy in real-world applications, is provided.

  \item The proposed system is validated in both simulation and real-world experiments.
        Our method outperforms many state-of-the-art visual(-inertial) localization or SLAM algorithms.
\end{itemize}

\section{Related Work}

In this section, we mainly discuss the literature focused on cross-modality localization, especially camera localization in 3D maps.

By finding the correspondences between two sensor inputs, several methods use common objects, which can be observed in both camera and lidar views.
In \cite{lu2017monocular}, the manually labeled road markings in a 3D lidar map were used to construct a sparse point cloud.
Combined with epipolar geometry and the vehicle odometry, the Chamfer distance between the edge image and the projected sparse point cloud was used to estimate 6-DoF camera poses.
In \cite{lu2017sharing}, vertical planes from both vision and lidar data, were extracted.
The authors took the correspondences of visual and lidar planes as coplanarity constraints to constrain the global bundle adjustment.

Mutual information is an effective metric for cross-modality matching, which is adopted in many methods to localize the camera in the maps produced by heterogenous sensors.
In \cite{wolcott2014visual}, the reflectivities from the lidar map were used to render synthetic images given the potential camera poses.
A 3-DoF search over the potential camera poses was applied.
The optimal pose was determined by maximizing the normalized mutual information between the camera input and the synthetic image to achieve 2D localization.
Using the derivatives of analytical normalized information distance (NID), Pascoe et al. \cite{pascoe2015robust} extended this method to 6-DoF camera pose estimation.
In \cite{pascoe2015farlap}, a similar NID-based method was proposed to evaluate the similarity between the live image and the images generated from a textured 3D prior mesh to obtain the camera pose.
Finally, based on mutual projections, the similarity between synthetic depth images and images from a panoramic camera was fit into a particle filter-based Monte Carlo localization framework in \cite{neubert2017sampling}.

Exploiting the geometric information is another strategy.
Typically, these methods will extract feature points in their visual modules, and are in the schemes of indirect visual methods.
Caselitz et al. \cite{caselitz2016monocular} introduced a monocular camera localization method which performs in an iterative closest point (ICP) scheme.
It associates and aligns the sparse point cloud produced by monocular visual SLAM with the lidar map iteratively to estimate the 7-DoF similarity transformation.
In \cite{kim2018stereo}, a method for stereo camera pose estimation was proposed.
It estimates the camera pose by minimizing the depth residuals between the depth from the stereo matching and the depth of the point projected from the lidar map.
Ding et al. \cite{ding2018laser} used a hybrid bundle adjustment to optimize the visual map from the stereo visual inertial system, and to align the sparse visual map against the pre-built lidar map at the same time.
Zuo et al. \cite{zuo2019visual} took the tightly-coupled MSCKF \cite{mourikis2007multi} as front-end tracking and registered the refined semi-dense point cloud from stereo matching to the prior lidar map using a normal distribution transform (NDT)-based method \cite{huhle2008registration}.
Using the Signed Distance Field (SDF) representation built from stereo vision, Huang et al. \cite{huang2019metric} proposed a monocular camera localization method by increasing the coherence between the indirect local structure and the SDF model.
Instead of using indirect visual pipelines, our method benefits from direct visual tracking \cite{engel2017direct}, which does not rely on the explicit feature extractors.
The correspondences of pixels among the frames can be updated during the optimization, and the plane information from the surfel representation can further help to make the system aware of the global pose and scale.

\section{Method}

\subsection{Notation}
In this paper, we denote the transformation matrix as \(\mathbf{T}^a_b \in \text{SE}(3)\), which transforms a point \(\mathbf{x}_b \in \mathbb{R}^3\) in the frame \(\mathcal{F}_b\) into the frame \(\mathcal{F}_a\).
The corresponding Lie-algebra elements \(\bm{\xi}^a_b \in \mathfrak{se}(3)\), which, for brevity, is expressed as a vector \(\bm{\xi}^a_b \in \mathbb{R}^6\), can be mapped by the exponential map, \(\exp(\bm{\xi}^a_b)\), to \(\mathbf{T}^a_b\).
The rotation matrix and translation vector of \(\mathbf{T}^a_b\) are denoted as \(\mathbf{R}^a_b \in \text{SO}(3)\) and \(\mathbf{t}^a_b \in \mathbb{R}^3\), respectively.
\(I_{(\cdot)}[\mathbf{p}]\) returns the pixel intensity of the image corresponding to the frame \(\mathcal{F}_{(\cdot)}\), given the homogeneous pixel coordinates \(\mathbf{p} = [u\ v\ 1]^\top\).
We use the pinhole model with \(\mathbf{K}\) as the camera intrinsic matrix and assume all images are undistorted.

\subsection{System Overview}

\begin{figure}[!ht]
  \centering
  \includegraphics[width=0.48\textwidth]{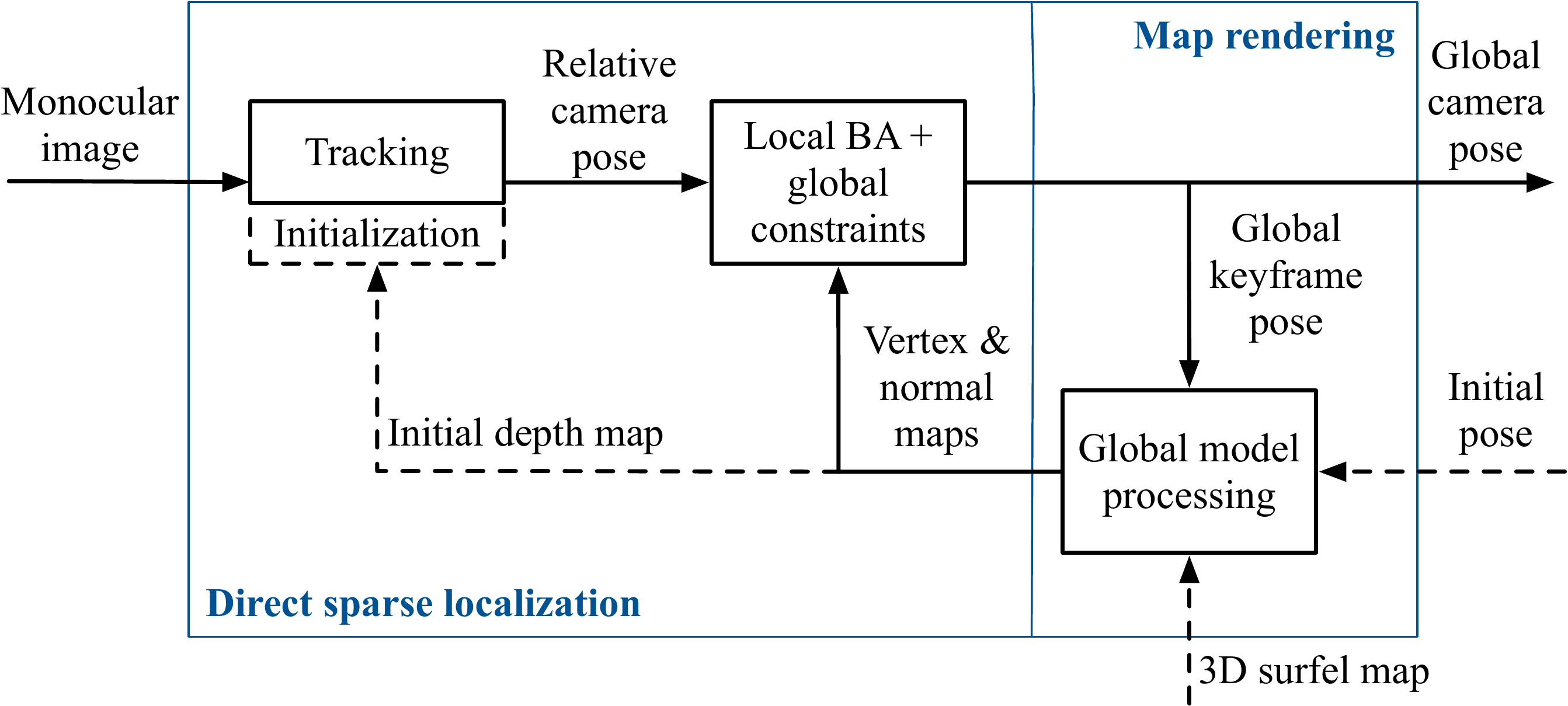}
  \caption{Framework of the proposed algorithm.
    Our system contains two sub-routines, the direct sparse localization and map rendering.
    Note that the parts represented by dashed lines in the diagram are only needed in the initialization.}
  \label{fig:system_diagram}
  \vspace{-1em}
\end{figure}

The framework of our method is shown in Fig.~\ref{fig:system_diagram}.
With a rough initial pose, our system first initializes direct sparse visual odometry with the generated depth map from the map rendering module (Sec. \ref{sec:map}).
With a valid value in the depth map, a candidate point is assigned a rough inverse depth, which is used for the future camera tracking.

\begin{figure}[t!]
  \centering
  \includegraphics[width=0.4\textwidth]{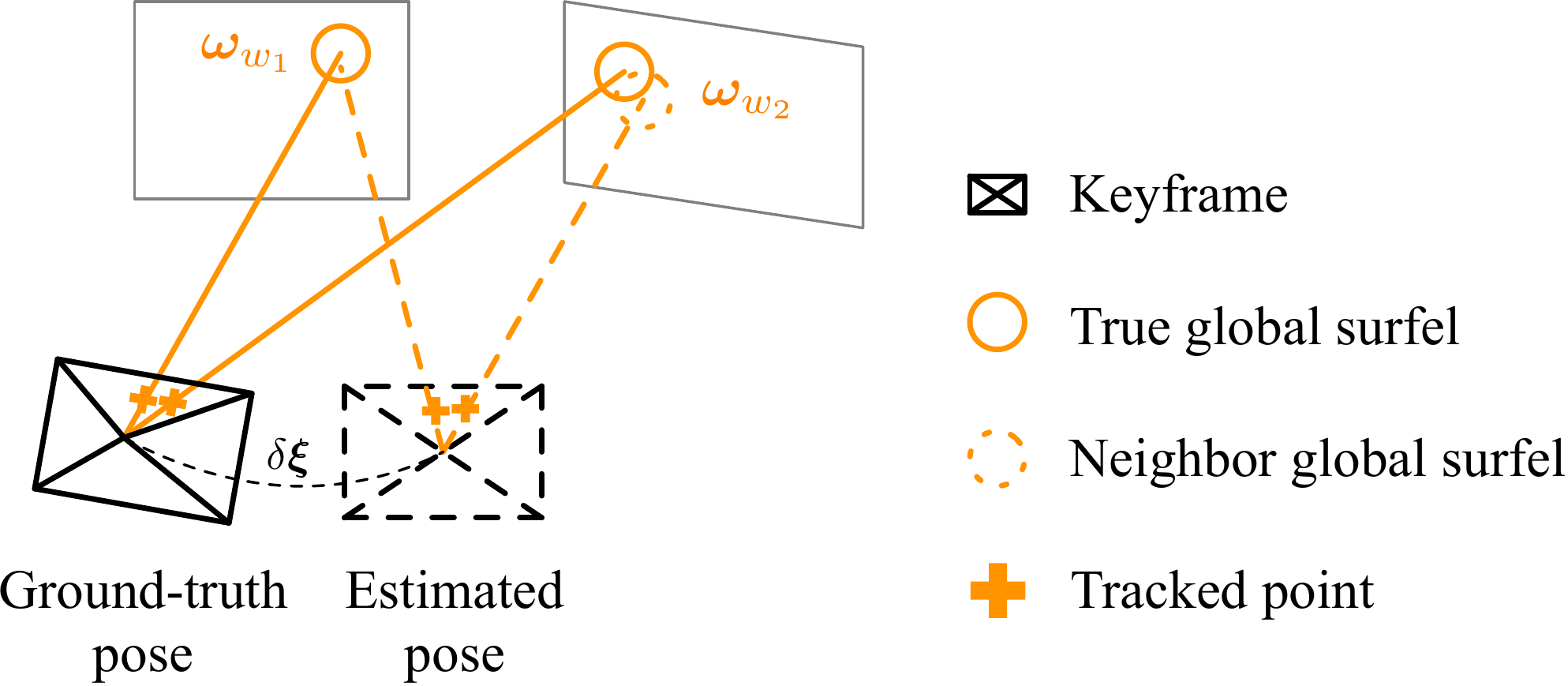}
  \caption{Assumption of global surfel association.
    If the error \(\delta \bm{\xi}\) between the ground-truth and the estimated poses is small, the tracked points can still be associated to the same global surfel or its neighbor surfels with close plane coefficients.}
  \label{fig:surfel_assumption}
\end{figure}

After initialization, the system obtains the vertex \& normal maps of the last keyframe, from the map rendering module (Sec. \ref{sec:map}).
This rendering step runs only once after a new keyframe is added and optimized.
In a local window with the number of keyframes \(N_\mathcal{F} = 7\), we track the points across all image regions following \cite{engel2017direct}.
We can acquire the plane information of the tracked points from the vertex and normal maps if one pixel is valid in both maps.
With the assumption that the local tracked points share the same plane in the world frame, we can ensure that most of the tracked points are associated with the correct global surfels, even though uncertainty may exist in the global keyframe pose.
This is illustrated in Fig. \ref{fig:surfel_assumption}.

Since each surfel can be considered as a local plane represented in the global frame, we use this plane information to constrain the relative poses between camera frames, as well as the global poses of frames hosting the sparse tracked points (Sec. \ref{sec:homo_constraint} and \ref{sec:opt}).

\subsection{Map Producing and Rendering} \label{sec:map}
Our surfel map is represented as a list of unordered surfels, similar to the ones proposed by Whelan et al. \cite{whelan2016elasticfusion}.
In our method, only the position, normal and radius are used for the camera pose estimation.

Provided a point cloud map from lidar mapping, we can build the surfel map as follows.
First, by voxel grid downsampling, we can reduce the number of points and make the points evenly distributed.
Then, the normal of each point is estimated by principal component analysis (PCA) of its neighbor points \cite{rusu2010semantic}.
The surfel map can be built by assigning each surfel by a point position, with its estimated normal in the processed point cloud and its radius according to the voxel size in the downsampling step.
When the system starts, this surfel map will be loaded once to GPU.

\begin{figure}[!htb]
  \centering
  \begin{subfigure}[b]{0.15\textwidth}
    \centering
    \includegraphics[width=\textwidth]{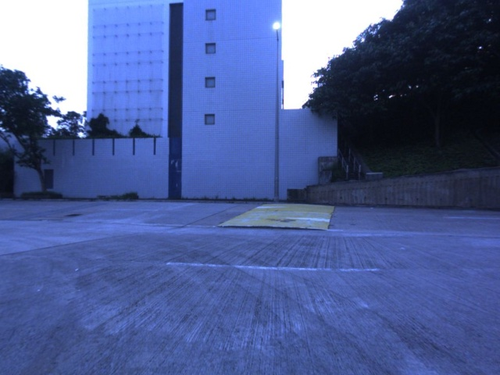}
    \caption{Raw input}
  \end{subfigure}
  \begin{subfigure}[b]{0.15\textwidth}
    \centering
    \includegraphics[width=\textwidth]{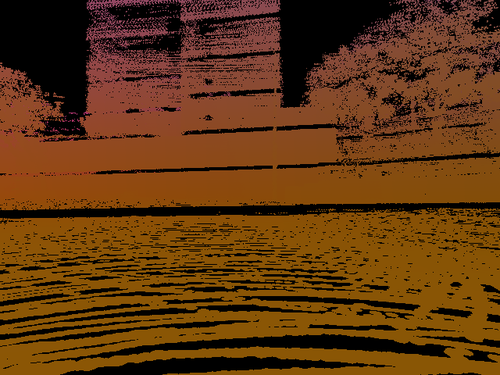}
    \caption{Vertex map \(\mathcal{M}_v\)}
  \end{subfigure}
  \begin{subfigure}[b]{0.15\textwidth}
    \centering
    \includegraphics[width=\textwidth]{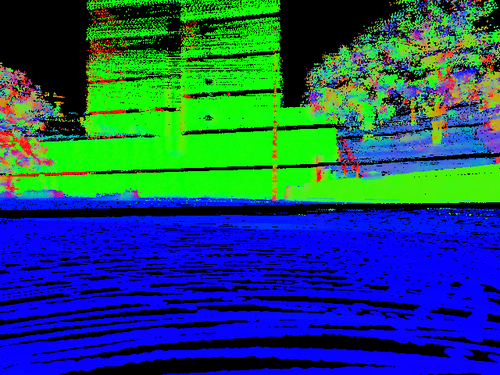}
    \caption{Normal map \(\mathcal{M}_n\)}
  \end{subfigure}
  \caption{A sample camera input and the rendered vertex \& normal maps given an estimated camera pose.
  The incompleteness of \(\mathcal{M}_{v,n}\) is due to the sparsity of the lidar inputs and different fields of view between the camera and lidar.
  }
  \label{fig:sample_input}
\end{figure}

Given a camera pose \(\mathbf{T}^w_c\) in the global frame, the map rendering module will project the surfels to the local frame \(\mathcal{F}_c\) and return the depth map, \(\mathcal{M}_d\), or vertex \& normal maps, \(\mathcal{M}_{v,n}\).
Similarly to \cite{whelan2016elasticfusion}, we use the OpenGL Shading Language to predict these maps.
The rendered maps have the same size as the input image.
For each pixel in the rendered maps, \(\mathcal{M}_d\) provides its depth in the given camera frame, while \(\mathcal{M}_{v,n}\) provides its surfel position and normal in the global frame.
Fig. \ref{fig:sample_input} shows a sample input image and the corresponding \(\mathcal{M}_{v,n}\) given an estimated camera pose.

\subsection{Homography Constraints from Global Surfel} \label{sec:homo_constraint}
To use the direct photometric errors as constraints, we follow \cite{engel2017direct} to formulate our energy functions.
For each tracked point, the photometric residual can be written as
\begin{equation} \label{eqn:photo_res}
  \begin{aligned}
    r^t_h =  I_t[\mathbf{p}_t] - \left(
    a^t_h \cdot I_h \left[ \mathbf{p}_h \right] + b^t_h
    \right)
  \end{aligned},
\end{equation}
where
\begin{equation}
  \begin{aligned}
    \begin{cases}
      a^t_h = 		\frac{t_t e^{a_t}}{t_h e^{a_h}},\ 
      b^t_h = b_t-\frac{t_t e^{a_t}}{t_h e^{a_h}} \cdot b_h \\
      \mathbf{p}_t \simeq \mathbf{K} \cdot (\mathbf{R}^t_h \bar{\mathbf{x}}_h + \mathbf{t}^t_h \rho_h)
    \end{cases}
  \end{aligned},
\end{equation}
where \(t_{(\cdot)}\) is the exposure time of the host or target image;
\(a_{(\cdot)}\) and \(b_{(\cdot)}\) are the coefficients for the affine brightness function \(e^{-a_{(\cdot)}}(I_{(\cdot)} - b_{(\cdot)})\);
and \(\rho_h\) is the inverse depth of the corresponding normalized point, \(\bar{\mathbf{x}}_h = \mathbf{K}^{-1} \cdot \mathbf{p}_h\), in the host frame, \(\mathcal{F}_h\); \(\simeq\) indicates equality up to a scale factor.

Given the plane coefficient of the pixel \(\mathbf{p}_h\) in \(\mathcal{F}_h\),
\({\bm{\omega}}_h =
\begin{bmatrix}
 {\mathbf{n}_h}^\top \ {d}_h
\end{bmatrix}^\top\),
so that for any point \(\mathbf{x}\) on the plane \({\mathbf{n}_h}^\top \mathbf{x} + d_h = 0\),
the homography \cite{hartley2003multiple} between \(\mathcal{F}_h\) and the target frame, \(\mathcal{F}_t\), can be written as
\begin{equation} \label{eqn:homo}
  \begin{aligned}
    \mathbf{H}^t_h & = \mathbf{K} \left(\mathbf{R}^t_h - \mathbf{t}^t_h \mathbf{n}_h^\top / d_h \right) \mathbf{K}^{-1} \\
    \mathbf{p}_t   & \simeq \mathbf{H}^t_h \mathbf{p}_h
  \end{aligned}.
\end{equation}

The variables to estimate are the relative poses \(\bm{\xi}^t_h\), and the affine brightness parameters \([a^t_h\ b^t_h]^\top\) between \(\mathcal{F}_h\) and \(\mathcal{F}_t\).
We denote the full variables as \(\mathcal{X}^t_h = [{\bm{\xi}^t_h}^\top \ a^t_h\ b^t_h]^\top \in \mathbb{R}^8\).
Note that \(\mathcal{M}_{v,n}\) stores vertex \& normal information in the global frame, \(\mathcal{F}_w\), from the map rendering module.
Thus, the plane information in \(\mathcal{F}_h\) needs to be transformed from the global frame with the estimated global pose of \(\mathcal{F}_h\), \( {\mathbf{T}}^h_w \), as
\begin{equation} \label{eqn:coeffs_transformation}
  \begin{aligned}
    \tilde{\bm{\omega}}_h =
    \begin{bmatrix}
      \tilde{\mathbf{n}}_h \\ \tilde{d}_h
    \end{bmatrix}
    =
    ({\mathbf{T}}^h_w)^{-\top} \cdot
    \begin{bmatrix}
      \mathbf{n}_w \\ d_w
    \end{bmatrix} = ({\mathbf{T}}^h_w)^{-\top} \cdot \bm{\omega}_w
  \end{aligned}.
\end{equation}

Combining Eqn. \ref{eqn:photo_res}, \ref{eqn:homo} and \ref{eqn:coeffs_transformation}, the photometric residual with the surfel constraint can be derived as
\begin{equation} \label{eqn:homo_res}
  \begin{aligned}
    r^t_h = &
    I_t \left[
        \mathbf{K} \cdot
        \left(\mathbf{R}^t_h - \mathbf{t}^t_h \tilde{\mathbf{n}}_h^\top / \tilde{d}_h
        \right) \mathbf{K}^{-1} \cdot \mathbf{p}_h
        \right] \\
            & -
    \left(
    a^t_h \cdot I_h \left[ \mathbf{p}_h \right] + b^t_h
    \right)
  \end{aligned}.
\end{equation}

Note that Eqn. \ref{eqn:homo_res} does not contain the inverse depths of the points with surfel constraints, but includes the global poses of the host frames, \(\mathbf{T}^h_w\).
This helps to constrain the camera poses globally in \(\mathcal{F}_w\).
For simplicity, we denote \(\bm{\xi}^{(\cdot)}_w\) as \(\bm{\xi}_{(\cdot)}\), and the to-be-optimized variables in \(\mathcal{F}_{(\cdot)}\) as \(\mathcal{X}_{(\cdot)} = [\bm{\xi}_{(\cdot)}^\top \ a_{(\cdot)}\ b_{(\cdot)}]^\top\).
Then, the Jacobian of \(r^t_h\) w.r.t. \(\mathcal{X}_h\) and \(\mathcal{X}_t\) can be written as
\begin{equation}
  \begin{aligned}
    \mathbf{J}^{r^t_h}_{\mathcal{X}_h} = \mathbf{J}^{r^t_h}_{\tilde{\mathbf{x}}_t} \cdot \mathbf{J}^{\tilde{\mathbf{x}}_t}_{\mathcal{X}_h} \\
    \mathbf{J}^{r^t_h}_{\mathcal{X}_t} = \mathbf{J}^{r^t_h}_{\tilde{\mathbf{x}}_t} \cdot \mathbf{J}^{\tilde{\mathbf{x}}_t}_{\mathcal{X}_t}
  \end{aligned},
\end{equation}
where
\begin{equation}
  \begin{aligned}
    \tilde{\mathbf{x}}_t
    =
    \left({\mathbf{R}}^t_h - {\mathbf{t}}^t_h \tilde{\mathbf{n}}_h^\top / \tilde{d}_h \right) \cdot \bar{\mathbf{x}}_h
  \end{aligned},
\end{equation}

\begin{equation}
  \begin{aligned}
    \mathbf{J}^{\tilde{\mathbf{x}}_t}_{\mathcal{X}_h} & = \frac{\partial \tilde{\mathbf{x}}_t(\bm{\omega}_h, \mathcal{X}^t_h)}{\partial \mathcal{X}_h}                                                                                                   \\
                                                   & \approx
    \frac{\partial \tilde{\mathbf{x}}_t}{\partial \bm{\omega}_h} \cdot \frac{\partial \bm{\omega}_h}{\partial \mathcal{X}_h} + \frac{\partial \tilde{\mathbf{x}}_t}{\partial \mathcal{X}^t_h} \cdot \frac{\partial \mathcal{X}^t_h}{\partial \mathcal{X}_h} \\
    \mathbf{J}^{\tilde{\mathbf{x}}_t}_{\mathcal{X}_t} & = \frac{\partial \tilde{\mathbf{x}}_t}{\partial \mathcal{X}^t_h} \cdot \frac{\partial \mathcal{X}^t_h}{\partial \mathcal{X}_t}
  \end{aligned}.
\end{equation}

\subsection{Optimization} \label{sec:opt}

\begin{figure}[t!]
  \centering
  \includegraphics[width=0.4\textwidth]{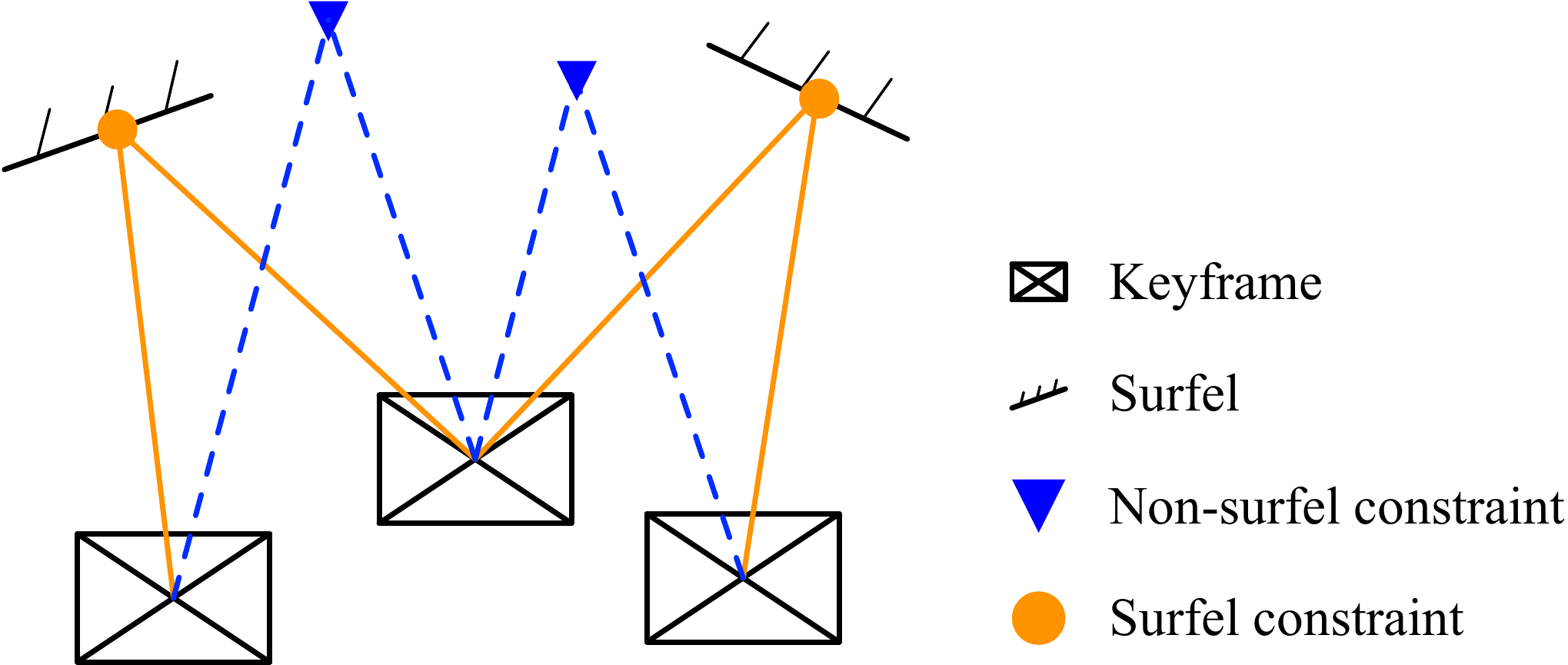}
  \caption{Surfel and non-surfel constraints.
  The point associated to a surfel forms the homography (surfel) constraints as Eqn. \ref{eqn:homo_res}, which constrains the global poses;
  the point without association forms normal photometric (non-surfel) constraints as Eqn. \ref{eqn:photo_res}, which constrains the relative pose between frames.}
  \label{fig:constraints}
\end{figure}

The final optimization is based on relative constraints from direct sparse tracking and global constraints from global surfels, as shown in Fig. \ref{fig:constraints}.
Due to map incompleteness or estimation uncertainty, not all tracked points can be associated with a global surfel.
Thus, the final energy function to be optimized becomes
\begin{equation} \label{eqn:joint_opt}
  \begin{aligned}
    E = E_\text{surfel} + E_\text{non}
  \end{aligned},
\end{equation}
where \(E_\text{surfel}\) and \(E_\text{non}\) are the energy function corresponding to the surfel and non-surfel constraints, respectively.
In detail,
\begin{equation}
  \begin{aligned}
    E_{(\cdot)} = \sum_{h \in \mathcal{K}}{\sum_{\mathbf{p} \in \mathcal{P}_h}{\sum_{t \in \text{obs}(\mathbf{p})}{\sum_{\mathbf{p}_h \in \mathcal{N}_{\mathbf{p}} }{\omega_r \left\Vert r^t_h \right\Vert_\gamma} } } }
  \end{aligned},
\end{equation}
where \(E_{(\cdot)}\) denotes \(E_\text{surfel}\) or \(E_\text{non}\), with the residual represented in Eqn. \ref{eqn:homo_res} or \ref{eqn:photo_res}, respectively;
\(\mathcal{K}\) is the set of all keyframes, \(\mathcal{P}_h\) is the set of tracked points (pixels) in \(\mathcal{F}_h\), \(\text{obs}(\mathbf{p})\) is the set of frames where the point \(\mathbf{p}\) is visible, and \(\mathcal{N}_\mathbf{p}\) is the pixels in the patch centered on \(\mathbf{p}\);
\(\omega_r\) is the gradient-dependent weighting defined in \cite{engel2017direct} and \(\left\Vert \cdot \right\Vert_\gamma\) is the Huber loss.
The above problem can be regarded as a non-linear least squares problem, which can be solved by the Gauss-Newton or Levenberg-Marquardt method.

\subsection{Implementation Details}
In this section, we briefly introduce the implementation details of the remaining parts of our system.

\subsubsection{Filtering and Association of Tracked Points}
We consider the pixels which have non-zero values on \(\mathcal{M}_{v,n}\) in the following.

After each iteration of the optimization, the updated inverse depth \(\tilde{\rho}_h\) and projected pixel coordinates in \(\mathcal{F}_t\), \(\tilde{\mathbf{p}}_t\), can be obtained.
From the intersection of the ray of the host point and the plane associated to the surfel, the inverse depth and projected pixel induced by the surfel can be obtained as \(\rho'_h\) and \(\mathbf{p}'_t\).
We \textit{filter} and \textit{associate} these tracked points by the following criteria:
\begin{itemize}
  \item If \(\left\Vert \tilde{\mathbf{p}}_t - \mathbf{p}'_t\right\Vert \geq 5\ \text{pixels}\) or \( \theta(\rho'_h, \tilde{\rho}_h) \geq 0.5\), the point is considered as an outlier and we will \textit{filter} it out,
  \item If \(\left\Vert \tilde{\mathbf{p}}_t - \mathbf{p}'_t\right\Vert < 2\ \text{pixels}\) and \( \theta(\rho'_h, \tilde{\rho}_h) < 0.2\), the point is regarded as a converged point and we \textit{associate} it to the corresponding surfel,
\end{itemize}
where \( \theta(\rho'_h, \tilde{\rho}_h) = 1 - [\min(\rho'_h, \tilde{\rho}_h) / \max(\rho'_h, \tilde{\rho}_h)] \).
The filtered point will be removed from Eqn. \ref{eqn:joint_opt},
while we will involve the associated point in \(E_\text{surfel}\) and remove it from \(E_\text{non}\).

After associating a point with a surfel, the inverse depth of the point will not be a variable to estimate and can be determined with the optimized pose of \(\mathcal{F}_h\).
Thus, the semi-dense depth map for frame tracking will set \(1 / \rho'_h\) as the pixel's depth, with the uncertainty obtained from the resolution of the map.

\subsubsection{Front-end}
For more details of the front-end implementation, we refer readers to \cite{engel2017direct}.
We follow similar frame and point management to that described in \cite{engel2017direct}.
The frame management tracks the frame by coarse-to-fine direct alignment, and creates and marginalizes keyframes to maintain the local window;
the point management selects, tracks and activates points for tracking and optimization.

\subsubsection{Marginalization}
For the points without surfel constraints, we can apply the same marginalization process as the one in \cite{engel2017direct}, where the First Estimate Jacobian \cite{huang2009first,leutenegger2015keyframe} is applied.
Since Eqn. \ref{eqn:homo_res} does not involve inverse depths, we do not need to marginalize these points with surfel constraints.
The related residuals of these points will be involved in the marginalization of frames only.

\section{Degeneration Analysis} \label{sec:analysis}
Degeneration appears when the surfel structure cannot constrain camera poses uniquely to the global surfel map.
In this section, several common degeneration cases are discussed.
Then, we demonstrate how the system recovers the localization drift with sufficient observations.
The detailed derivations for this section can be found in the supplementary material \cite{ye2020supplementary}.

Surfel distributions, especially the normal directions of surfels, can influence the performance of the proposed system.
The tracked points with no surfel associations can provide relative constraints for camera poses.
If all the constraints are from non-surfel points, the system is equivalent to visual-only odometry, which has scale ambiguity \cite{jones2011visual}.

To simplify the analysis, we consider that the constraints are directly from points in the image planes \(\bar{\mathbf{x}}_t\) and \(\bar{\mathbf{x}}_h\), instead of their intensity values, and that the pose of the first frame is given.
Then, in this visual-only case, the camera poses \( \mathcal{F}_t \) and \( \mathcal{F}_h \) are constrained by the following relationship:
\begin{equation} \label{eqn:projection}
  \begin{aligned}
    \bar{\mathbf{x}}_t \simeq \left( \mathbf{R}^t_h \cdot \bar{\mathbf{x}}_h + \mathbf{t}^t_h \cdot  \rho_h \right)
  \end{aligned}.
\end{equation}

For any scale factor \(\lambda\) and global transform \(\bar{\mathbf{T}}\) (with corresponding rotation \(\bar{\mathbf{R}}\) and translation \(\bar{\mathbf{t}}\)), identical measurements of \(\bar{\mathbf{x}}_h\) and \(\bar{\mathbf{x}}_t\) are produced by the following variables with tilde sign
\begin{equation} \label{eqn:ambiguity}
  \begin{aligned}
    \begin{cases}
      \tilde{\mathbf{R}}^w_h = \bar{\mathbf{R}} \mathbf{R}^w_h,\
      \tilde{\mathbf{R}}^w_t = \bar{\mathbf{R}} \mathbf{R}^w_t                              \\
      \tilde{\mathbf{t}}^w_h = \lambda (\bar{\mathbf{R}} \mathbf{t}^w_h + \bar{\mathbf{t}}),\
      \tilde{\mathbf{t}}^w_t = \lambda (\bar{\mathbf{R}} \mathbf{t}^w_t + \bar{\mathbf{t}}) \\
      \tilde{\rho}_h = \rho_h / \lambda
    \end{cases}
  \end{aligned}.
\end{equation}
By Eqn. \ref{eqn:ambiguity}, the relative pose between \( \mathcal{F}_t \) and \( \mathcal{F}_h \) and the inverse depth of any point from non-surfel constraints become
\begin{equation} \label{eqn:ambiguity_relative}
  \begin{aligned}
      \tilde{\mathbf{R}}^t_h =  \mathbf{R}^t_h,\
      \tilde{\mathbf{t}}^t_h =  \lambda \mathbf{t}^t_h,\
      \tilde{\rho}_h         = \rho_h / \lambda
  \end{aligned}.
\end{equation}

The identical measurements can be verified by substituting Eqn. \ref{eqn:ambiguity_relative} into Eqn. \ref{eqn:projection} as
\begin{equation} \label{eqn:rel_pose}
  \begin{aligned}
    \bar{\mathbf{x}}_t \sim \left( \tilde{\mathbf{R}}^t_h \cdot \bar{\mathbf{x}}_h + \tilde{\mathbf{t}}^t_h \cdot \tilde{\rho}_h \right) = \left( \mathbf{R}^t_h \cdot \bar{\mathbf{x}}_h + \mathbf{t}^t_h \cdot \rho_h \right)
  \end{aligned}.
\end{equation}

To analyze the degeneration with surfel constraints, we make two assumptions:
1) The first assumption is that the visual system can track points ideally.
This leads to a relatively accurate visual structure, camera poses up to scale and an unknown global transform.
2) The second assumption is that the surfel coefficients, as well as the associations between tracked points and surfels, are known.

From the above assumptions and the accurate relative pose relationship from Eqn. \ref{eqn:rel_pose}, we can regard \(\tilde{\mathbf{R}}^t_h\), \(\tilde{\mathbf{t}}^t_h\), \(\tilde{\rho}_h\), \(\bar{\mathbf{x}}_h\), \(\bar{\mathbf{x}}_t\) and \(\bm{\omega}_w\) as known and locally constrained.
The uncertainty comes from the unknown scale \(\lambda\) and global transform \(\bar{\mathbf{T}}\).

We can rewrite the surfel constraints as
\begin{equation} \label{eqn:surfel_exp}
  \begin{aligned}
     & \bar{\mathbf{x}}_t \simeq \left( \tilde{\mathbf{R}}^t_h \cdot \bar{\mathbf{x}}_h + \tilde{\mathbf{t}}^t_h \cdot \rho'_h \right),                                        \\
     & \text{where}\ \rho'_h = -\frac{ {\mathbf{n}_w}^{\top} \cdot {\tilde{\mathbf{R}}_h^w} }{{{}\tilde{\mathbf{t}}^w_h}^\top \cdot \mathbf{n}_w + d_w} \cdot\bar {\mathbf{x}}_h
  \end{aligned}.
\end{equation}

The degeneration exists when two or more different state pairs (\(\mathbf{T}\) and \(\rho\)) hold the same constraints from Eqn. \ref{eqn:surfel_exp}.

\subsection{Single Plane} \label{sec:single_plane}
The first degeneration case is when all points are on the same plane and they share the same plane coefficients, \( \bm{\omega}_w \).
In this case, the identical measurements can be produced by
\begin{equation}
  \begin{aligned}
      \bar{\mathbf{R}}^\top \cdot \mathbf{n}_w = \mathbf{n}_w,\
      \bar{\mathbf{t}}^\top \cdot \mathbf{n}_w = (\lambda - 1) \cdot d_w
  \end{aligned}.
\end{equation}
Neither \(\lambda\) nor \(\bar{\mathbf{T}}\) can be uniquely determined.

\subsection{Parallel Planes} \label{sec:parallel_planes}
The parallel planes case will appear when all the points are from two sides of a long passage.
In this case, the absolute scale \(\lambda\) can be determined.
However, the global transform \(\bar{\mathbf{T}}\) still remains distinguished.
Any \(\bar{\mathbf{R}}\) and \(\bar{\mathbf{t}}\) meeting
\begin{equation}
  \begin{aligned}
      \bar{\mathbf{R}}^\top \cdot \mathbf{n}_w = \mathbf{n}_w,\
      \bar{\mathbf{t}}^\top \cdot \mathbf{n}_w = 0
  \end{aligned}
\end{equation}
will not violate the surfel or non-surfel constraints.
It can be considered as a particular case of Sec. \ref{sec:single_plane}, where the scale \(\lambda = 1\). \(\bar{\mathbf{T}}\) can be formed by any rotation aligned with the plane normal, \(\mathbf{n}_w\), and any translation perpendicular to \(\mathbf{n}_w\).

\subsection{Non-parallel Planes with Co-planar Normals} \label{sec:co_planar}
In this case, all normal vectors of the surfel constraints spread on a plane, \(\Omega\), only.
We denote the normal vector of \(\Omega\) as \(\mathbf{n}_\Omega\).
Different from the case in Sec. \ref{sec:parallel_planes}, two or more non-parallel planes now exist.
There will be no ambiguity on the global rotation and scale, i.e., \(\bar{\mathbf{R}} = \mathbf{I}_{3 \times 3}\) and \(\lambda = 1\).
The ambiguity appears only when \(\bar{\mathbf{t}}\) satisfies \(\mathbf{n}_w^\top \bar{\mathbf{x}} (\bar{\mathbf{t}}^\top \mathbf{n}_w) = 0\) for all \(\bar{\mathbf{x}}\) and \(\mathbf{n}_w\).
Since all normal vectors spread on \(\Omega\), any \(\bar{\mathbf{t}}\) meeting
\begin{equation}
  \begin{aligned}
    \bar{\mathbf{t}} \parallel \mathbf{n}_\Omega
  \end{aligned}
\end{equation}
leads to \(\bar{\mathbf{t}}^\top \mathbf{n}_w = 0\).
Thus, \(\bar{\mathbf{T}}\) cannot be determined uniquely.

\subsection{Recovery with Sufficient Observations} \label{sec:sufficient_obs}

If there are sufficient and diverse surfels (\(\mathbf{n}_w\) and \(d_w\)) observed in multiple host frames, \(\lambda\) and \(\bar{\mathbf{T}}\) can be determined uniquely.
The influence of the surfel distribution on the localization accuracy is further evaluated by simulation in Sec. \ref{sec:simulation}.
\section{Results}
In this section, quantitative results are provided to validate our method.
Fig. \ref{fig:to_ias} shows some qualitative results on our HKUST dataset with lidar, IMU, camera and GPS data collected from a golf cart.
More qualitative results can be found in our supplementary material \cite{ye2020supplementary} and \href{https://youtu.be/LTihCBGcURo}{video}.

\subsection{EuRoC Indoor Quantitative Results}

We compared our DSL method with the state-of-the-art stereo-inertial localization method (MSCKF w/ map) \cite{zuo2019visual}, the visual-inertial SLAM method with loop-closures (VINS-Mono) \cite{qin2018vins} and direct sparse odometry (DSO) \cite{engel2017direct} on the EuRoC dataset \cite{burri2016euroc}.
The EuRoC dataset provides stereo grayscale images, IMU data, ground-truth poses and a ground-truth lidar map.
For the following results, our method and DSO\footnote{The camera inputs for DSL and DSO are photometricaly calibrated by \cite{bergmann2017online} for the V1\_03\_difficult to compensate for the unknown exposure time.} were evaluated on one of the cameras as inputs only;
Both camera inputs and IMU data are used for MSCKF w/ map;
And for VINS-Mono, the left camera and IMU data are inputs.

When our localization method starts, the initial pose of the first camera frame is provided.
We found that our system could recover from the initial guess with perturbations around 0.3 m and 5 degrees, thanks to the constraints from the global model.

The absolute trajectory error (ATE) of each sequence and relative pose error (RPE) over all trajectories \cite{sturm2012benchmark} are shown in Table \ref{tab:euroc_results}, where the results of MSCKF w/ Map and VINS-Mono (loop) are reported by \cite{zuo2019visual}.
The estimated and ground-truth poses were aligned for all methods and scaled for the monocular method DSO by \cite{umeyama1991least}.
The results were averaged over 5 runs to reduce the randomness.
In the ATE results, our method outperforms the visual-only and visual-inertial methods.
In the RPE results, our method has close results w.r.t. different lengths of the trajectory segment.
This shows that our method can provide both short- or long-distance pose estimation accurately.

\begin{table*}[!ht]
  \centering
  \caption{Average ATE and RPE \cite{sturm2012benchmark} for 5 runs are shown in the left and right tables, respectively.
    Units are in meters.}
  \begin{tabular}{cc}
    \begin{tabular}{lC{0.65cm}C{0.65cm}C{0.65cm}C{0.65cm}cc}
      \toprule
      \multicolumn{1}{l}{\multirow{2}[2]{*}{\textbf{Dataset}}} & \multicolumn{2}{c}{\textbf{DSL (ours)}} & \multicolumn{2}{c}{\textbf{DSO}} & \multicolumn{1}{c}{\multirow{2}[2]{*}{ \shortstack{ \textbf{MSCKF}                         \\ \textbf{w/ Map} }}} & \multicolumn{1}{c}{\multirow{2}[2]{*}{ \shortstack{ \textbf{VINS-Mono}\\ \textbf{(loop)} } }} \\
      \cmidrule(lr){2-3}\cmidrule(lr){4-5}                     & left                                    & right                            & left                                                               & right &       &       \\
      \midrule
      V1\_01\_easy                                             & \textbf{0.035}                          & \textbf{0.039}                   & 0.091                                                              & 0.065 & 0.056 & 0.044 \\
      V1\_02\_medium                                           & \textbf{0.034}                          & \textbf{0.026}                   & 0.212                                                              & 0.177 & 0.055 & 0.054 \\
      V1\_03\_difficult                                        & \textbf{0.045}                          & \textbf{0.047}                   & 0.161                                                              & 0.234 & 0.087 & 0.209 \\
      V2\_01\_easy                                             & \textbf{0.026}                          & \textbf{0.023}                   & 0.047                                                              & 0.043 & 0.069 & 0.062 \\
      V2\_02\_medium                                           & \textbf{0.023}                          & \textbf{0.025}                   & 0.074                                                              & 0.08  & 0.089 & 0.114 \\
      V2\_03\_difficult                                        & \textbf{0.103}                          & \textbf{0.083}                   & X                                                                  & X     & 0.149 & 0.149 \\
      \bottomrule
    \end{tabular}%
    \label{tab:euroc_ape}%
     &
    \begin{tabular}{lC{0.65cm}C{0.65cm}cc}
      \toprule
      \multicolumn{1}{l}{\multirow{2}[2]{*}{\shortstack[l]{\textbf{Segment} \\ \textbf{Length}} } } & \multicolumn{2}{c}{\textbf{DSL (ours)}} & \multicolumn{1}{c}{\multirow{2}[2]{*}{\shortstack{ \textbf{MSCKF} \\ \textbf{w/ Map} }} } & \multicolumn{1}{c}{\multirow{2}[2]{*}{\shortstack{ \textbf{VINS-Mono}\\ \textbf{(loop)}} }} \\
      \cmidrule{2-3} & left           & right          &       &            \\
      \midrule
      7m             & \textbf{0.121} & \textbf{0.111} & 0.143 & 0.156      \\
      14m            & \textbf{0.121} & \textbf{0.106} & 0.154 & 0.160      \\
      21m            & \textbf{0.133} & \textbf{0.120} & 0.184 & 0.208      \\
      28m            & \textbf{0.108} & \textbf{0.100} & 0.175 & 0.223      \\
      35m            & \textbf{0.118} & \textbf{0.111} & 0.191 & 0.260      \\
      \bottomrule
    \end{tabular}%
    \label{tab:euroc_rpe}%
  \end{tabular} \label{tab:euroc_results}
  \vspace{-1em}
\end{table*}

\subsection{CARLA Simulator Outdoor Tests} \label{sec:simulation}

\begin{figure}
  \centering
  \begin{subfigure}[b]{0.22\textwidth}
    \centering
    \includegraphics[width=\textwidth]{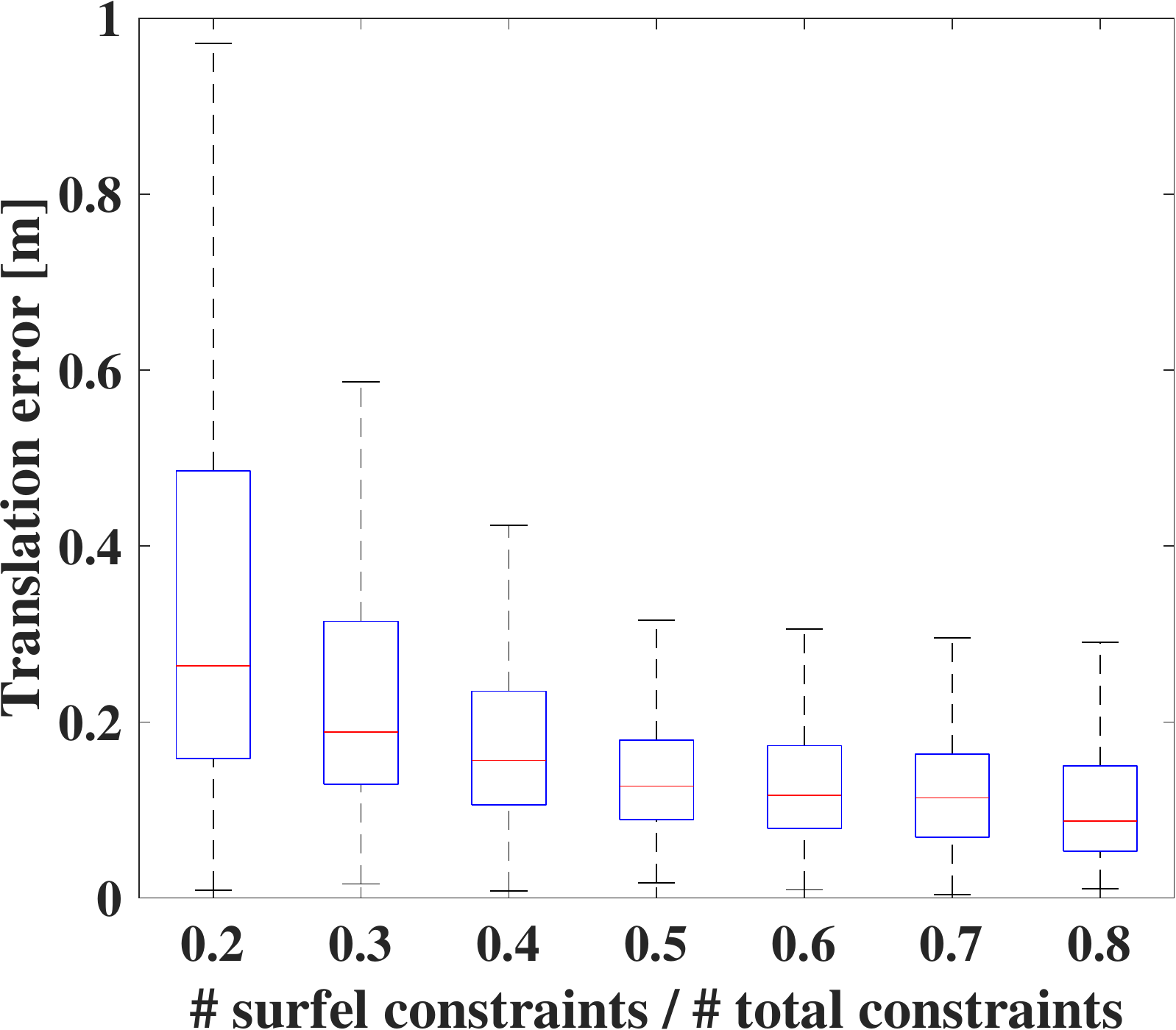}
    \caption{Ratio of constraints}
    \label{fig:surfel_ratio}
  \end{subfigure}
  \begin{subfigure}[b]{0.25\textwidth}
    \centering
    \includegraphics[width=\textwidth]{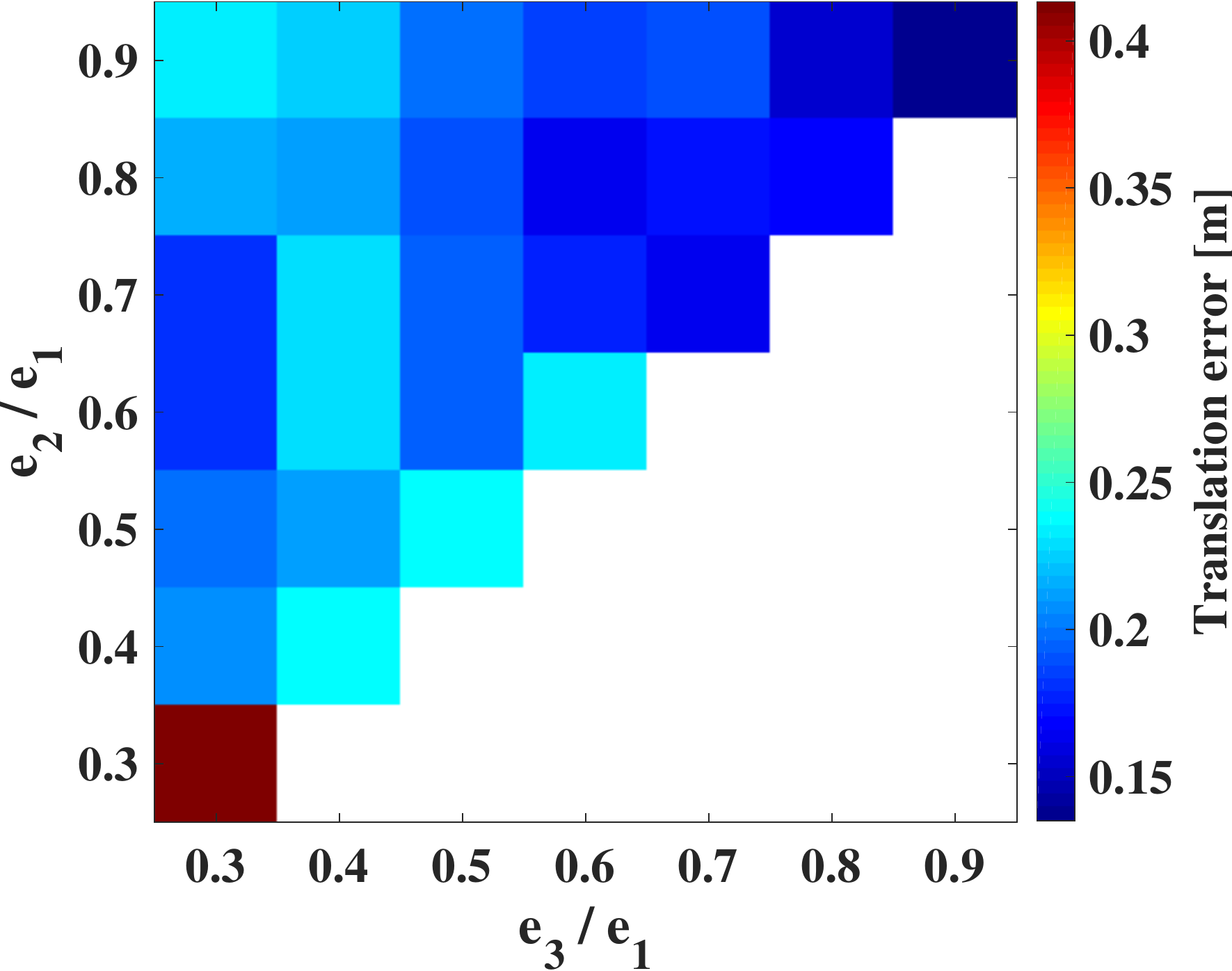}
    \caption{Surfel distribution}
    \label{fig:surfel_normals}
  \end{subfigure}
  \caption{Localization performance under different surfel conditions (the x-axis of (a) and two axes of (b) denote the ratio within a domain, e.g., \(0.3\) represents \((0.2, 0.3]\)). (a) Translation errors of camera poses w.r.t. the ratio of the surfel constraints to the total constraints. (b)  Translation errors w.r.t. the ratio of eigenvalues of the covariance matrix of the surfel normals, \(\text{e}_i\) (better viewed in color).}
  \label{fig:surfel_analysis}
\end{figure}

We next evaluated our method within the CARLA simulator \cite{dosovitskiy2017carla}, which is capable of generating maps, camera inputs and ground-truth poses.
To evaluate the effects of surfel distribution and the ratio of the surfel constraints to the total constraints, we collected the localization errors and all the constraints used at the same time.

Due to the estimation errors of the inverse depths and incompleteness in the rendered maps \(\mathcal{M}_{v,n}\), as similarly shown in Fig. \ref{fig:sample_input}, not all pixels can be associated to surfels.
We tested our method with different randomly sampled maps.
In Fig. \ref{fig:surfel_ratio}, the translation errors of camera poses w.r.t. the ratio of the surfel constraints to the total constraints are shown.
We can see that large errors exist when the surfel constraints are not sufficient, i.e., the ratio \(\leq 0.2\).
This is because when there are insufficient surfel constraints, our method degrades to a monocular visual method, which can have scale- or pose- drift without the global constraints.
Thus, to ensure accurate results, a surfel map covering most of the camera observations is recommended.
In practice, the lidar map used to produce the surfel map should have sufficient overlap with the camera inputs.

To show the effects of surfel distribution, we collected the plane coefficients of the surfel constraints in each frame.
We could obtain the covariance matrix of all surfel normals, whose eigenvalues are denoted as \(\text{e}_i, i = \{1,2,3\}\), where \(\text{e}_1 > \text{e}_2 > \text{e}_3\).
The ratios of \(\text{e}_i\) were calculated, and are compared with the errors of the camera poses in Fig. \ref{fig:surfel_normals}.
We can see that on the bottom left of the figure, the translation errors are larger because all the surfels have almost the same normal direction, corresponding to the case in Sec. \ref{sec:single_plane} or \ref{sec:parallel_planes},
while on the top right of the figure, the error is smaller, where surfel normals are distributed evenly in space.
These results are consistent with our analysis in Sec. \ref{sec:analysis}.

Furthermore, to show the influence of map noises, we added Gaussian noise with different noise levels \(\sigma\) to the original point cloud and re-generated the surfel maps.
The translation and rotation errors w.r.t. different map noises are shown in Table \ref{tab:map_noises}.
We found that our method could still have reliable localization performance with standard deviation \(\sigma = 0.4\text{m}\).
When the noise was too large, the pose accuracy degraded due to the inaccuracy of the the normal estimation and of the position of the vertices.
However, this extreme case could be avoided by checking the map quality.

\begin{table}[!htbp]
  \centering
  \caption{Average pose errors of DSL w.r.t. map noises.}
  \begin{tabular}{lcccccc}
    \toprule
    \textbf{Surfel noise} \(\sigma\) [m] & {0.0} & 0.1  & 0.2  & 0.3  & 0.4  & 0.5  \\
    \midrule
    Translation error [m]                & 0.12  & 0.19 & 0.22 & 0.35 & 0.52 & 2.08 \\
    Rotation error [deg]                 & 0.29  & 0.57 & 0.49 & 0.74 & 0.86 & 0.88 \\
    \bottomrule
  \end{tabular}%
  \label{tab:map_noises}%
  \vspace{-1.5em}
\end{table}%

\subsection{Runtime}

Runtime analysis\footnote{Run on an Intel i7-8700K CPU with an Nvidia GTX-1080Ti GPU.} on different datasets can be found in Table \ref{tab:runtime}.
Compared to the runtime of DSO \cite{engel2017direct}, our proposed method had almost no additional overhead by involving the global surfel constraints and rendering.

\begin{table}[htbp]
  \centering
  \caption{Runtime analysis on different datasets.}
  \begin{tabular}{lccc}
    \toprule
    \textbf{Dataset}  & \textbf{EuRoC}     & \textbf{CARLA}     & \textbf{HKUST}    \\
    \midrule
    Rendering (ms)    & 8 \(\pm\) 1        & 11 \(\pm\) 1       & 9 \(\pm\) 1          \\
    Tracking (ms)     & 22 \(\pm\) 20      & 23 \(\pm\) 19      & 15 \(\pm\) 7         \\
    Optimization (ms) & 117 \(\pm\) 31     & 112 \(\pm\) 37     & 102 \(\pm\) 34       \\
    Number of surfels & 3.20E+06           & 8.24E+06           & 9.44E+06             \\
    Surfel radius (m) & 0.01               & 0.1                & 0.05                 \\
    Image size        & 752 \(\times\) 480 & 800 \(\times\) 600 & 640   \(\times\) 480 \\
    \bottomrule
  \end{tabular}%
  \label{tab:runtime}%
  \vspace{-1.5em}
\end{table}%

\section{Conclusion} \label{sec:conclusion}
In this work, we have introduced a cross-modality algorithm of monocular direct sparse camera localization in a prior surfel map (DSL), which has the ability to provide accurate 6-DoF camera poses.
The proposed method uses surfel representation of the 3D map.
Given an estimated pose, we render the surfels into vertex and normal maps, from which we obtain the plane coefficients of the associated pixels.
The plane coefficients of the surfels form the proposed homography constraints to make the whole system aware of the absolute scale and global poses.
The final optimization combines the tracked points with and without surfel constraints in a fully direct photometric formulation.
We have also shown the degeneration analysis of our method, which can be used to indicate the reliability of the system.
Comprehensive evaluation shows that our method outperforms many state-of-the-art visual(-inertial) localization or SLAM algorithms.
Our future work will investigate the possibility of online map updating by camera observations and applying DSL in more dynamic and challenging scenarios.

\addtolength{\textheight}{-3.65cm}   %

\bibliographystyle{IEEEtran}
\bibliography{root_bib}

\end{document}